%% file: name.tex
\documentclass[conference,a4paper]{IEEEtran}

\ifCLASSINFOpdf
 
\else
 
\fi

\usepackage{graphicx}
\usepackage{amsmath}
\usepackage{amssymb}
\usepackage{booktabs}

\usepackage{multirow}
\usepackage{color}
\usepackage{stfloats}
\usepackage[table,xcdraw]{xcolor}
\usepackage{etoolbox}

\usepackage[pagebackref,breaklinks,colorlinks]{hyperref}

\usepackage[capitalize]{cleveref}
\crefname{section}{Sec.}{Secs.}
\Crefname{section}{Section}{Sections}
\Crefname{table}{Table}{Tables}
\crefname{table}{Tab.}{Tabs.}

\begin{document}

\title{Fully Aligned Network for Referring \\Image Segmentation }
% \title{Reasonable Interaction Designs Boost Referring Image Segmentation}

\author{Yong Liu\textsuperscript{1}~,
        Ruihao Xu\textsuperscript{1}~,
        Yansong Tang\textsuperscript{1}\\
        % {\tt\small liuyong23@mails.tsinghua.edu.cn, tang.yansong@sz.tsinghua.edu.cn}\\
\textsuperscript{1}Tsinghua Shenzhen International Graduate School, Tsinghua University}
% \textsuperscript{2}ByteDance Inc.~
% \textsuperscript{3}The University of Hong Kong\\
% {\tt\small liuyong23@mails.tsinghua.edu.cn, tang.yansong@sz.tsinghua.edu.cn}
% }

\maketitle
%%%%%%%%% ABSTRACT
\begin{abstract}

This paper focuses on the Referring Image Segmentation (RIS) task, which aims to segment objects from an image based on a given language description. 
The critical problem of RIS is achieving fine-grained alignment between different modalities to recognize and segment the target object. 
Recent advances using the attention mechanism for cross-modal interaction have achieved excellent progress. However, current methods tend to lack explicit principles of interaction design as guidelines, leading to inadequate cross-modal comprehension. Additionally, most previous works use a single-modal mask decoder for prediction, losing the advantage of full cross-modal alignment. To address these challenges, we present a Fully Aligned Network (FAN) that follows four cross-modal interaction principles. Under the guidance of reasonable rules, our FAN achieves state-of-the-art performance on the prevalent RIS benchmarks (RefCOCO, RefCOCO+, G-Ref)  with a simple architecture.
\end{abstract}

\section{Introduction}
Referring Image Segmentation (RIS) \cite{RIS1,rrn} aims to segment the target object in an image based on a given text description. RIS requires understanding the content of different modalities to identify and segment the target accurately. This task is crucial in multi-modal research~\cite{soc,gkc,scan,unilseg}, with applications in human-robot interaction and image processing \cite{qdmn,gsfm,quality,morn,uvcom}.

The main challenge in RIS is aligning different modalities due to varied image content and unrestricted language expression. Early methods \cite{rrn,dmn} concatenated linguistic features with vision features but performed poorly due to lack of cross-modal interaction. Later methods \cite{busnet,cmpc} used multi-modal graph reasoning to localize referred objects based on detailed descriptions. With the development of transformer~\cite{transformer,riformer,fvstream}, taking cross-attention operation for vision and language alignment has received growing interest~\cite{vlt, lavt, cris}. 
% However, reliance on single-modal mask decoders and lack of explicit interaction principles hinder further progress.
However, there remain two potential problems that constrain the development of this field.
Firstly, almost all current methods take a single-modal mask decoder to output the prediction mask. Due to the lack of vision-and-language interaction, the mask decoder tends to lose the advantage of fully utilizing multi-modal guidance.
Secondly, the design of previous models lacks explicit alignment principles as guidance, which may lead to insufficient cross-modal alignment.
As a result, they usually design respective auxiliary modules to improve performance. But these auxiliary modules are often not generalizable.

To this end, we summarize four cross-modal interaction principles and present a simple, clean yet strong Fully Aligned Network (FAN). 
The structure design of FAN is guided by the following principles:
\textit{Encoding Interaction:} performing preliminary activation of visual features, which helps to alleviate the effect of background pixels. \textit{Coarse and Fine-Grained Interaction:} utilizing both word-level and sentence-level features for detailed target object highlighting.
\textit{Multi-Scale Interaction:} leveraging diverse information from visual features at hierarchical scales. 
\textit{Bidirectional Interaction:} updating visual and linguistic features simultaneously to create a joint space by producing implicit content-aware expressions that are more suitable for model understanding.

With these principles, FAN builds a well-aligned visual and textual common space using attention operations, which allows the prediction mask can be generated by simple similarity calculation without the need for a complex operation. Our experiments on RefCOCO~\cite{refcoco}, RefCOCO+\cite{refcoco}, and G-Ref~\cite{grefcoco} datasets show that FAN achieves excellent performance. Our contributions can be concluded as follows:
\begin{itemize}
    % \item We summarize four interaction principles that help to build deep cross-modal relationships between image content and language description. 
    % TODO：是否要突出原则有助于fully aligned的概念？
    \item We propose explicit interaction principles that help to build deep cross-modal relationships between image content and language description. Guided by that, we design a conceptually simple, clean, yet strong framework named Fully Aligned Network (FAN), which achieves fully cross-modal alignment with a attention mechanism.
    \item Our FAN achieves state-of-the-art performance on the popular dataset: RefCOCO, RefCOCO+, and G-Ref. 
\end{itemize}

\section{Related Work}
Referring image segmentation (RIS) segments pixels into masks based on natural language expressions, requiring effective cross-modal alignment. Initial baselines include \cite{early-work1}, \cite{early-work2}. Subsequent methods generally fall into two main categories.

The first idea is to utilize text structure to excavate linguistic relationships further for object targeting.
MAttNet~\cite{mattnet} proposes to decompose the description into different modular components related to appearance, location, and relationships.
Some other methods~\cite{lscm, busnet, cmpc} leverage the graph networks to model the internal structure of the text.
However, the above methods do not model well-aligned cross-modal common space, and their pipelines tend to be complex.

The other idea is to model the cross-modal relations between image and language by various attention operations.
% STEP~\cite{step} groups pixels by calculating the similarity between the transformed linguistic and visual embedding.
KWAN~\cite{kwn} utilizes the cross-modal cross-attention to build the joint space.
EFN~\cite{efn} and LAVT~\cite{lavt} propose to fuse inside the visual backbone.
CRIS~\cite{cris} leverages the CLIP~\cite{CLIP} pre-trained weights with a contrastive loss.

Recent advances~\cite{step, lavt, unilseg} have achieved excellent performance but lack explicit alignment principles. Additionally, most previous works use a single-modal mask decoder for prediction, which misses the benefits of full cross-modal alignment. To this end, we propose explicit interaction principles and introduce a conceptually simple, clean, yet strong framework called the Fully Aligned Network (FAN).

\begin{figure*}[t]
    \centering
    \includegraphics[width=0.85\textwidth]{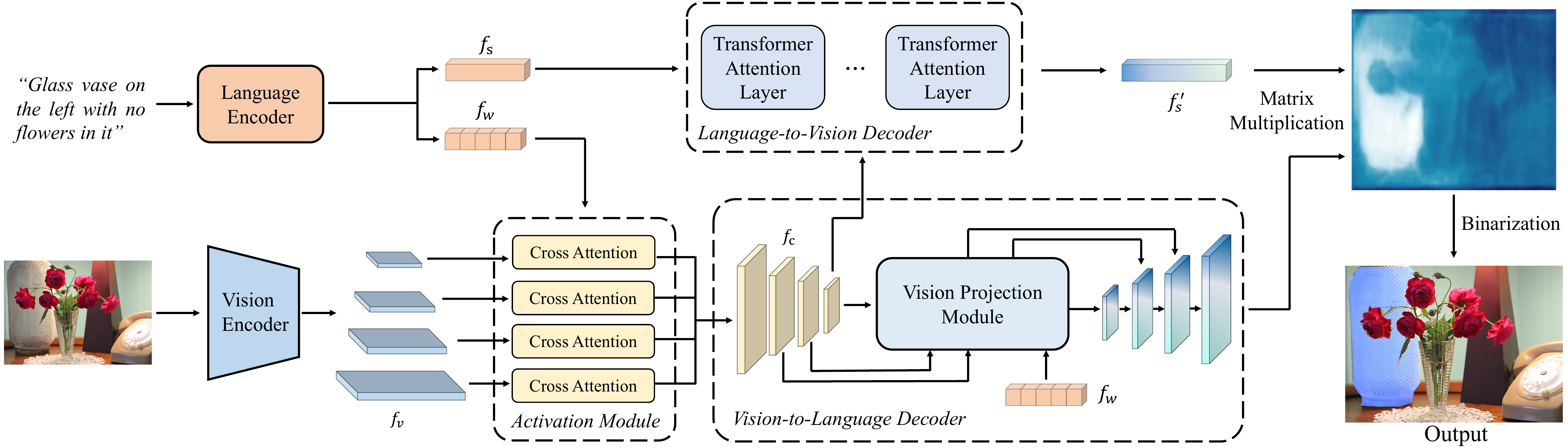}
    \vspace{-5pt}
    \caption{Pipeline of our FAN. Taking an image and the corresponding language expression as input, the vision and language encoder extract corresponding features, respectively. Then a multi-scale activation module performs preliminary fusion between them to highlight the referred region roughly. For the decoding process, we update visual and linguistic features simultaneously to project them into the common space. Finally, the output mask is obtained by simple similarity calculation and binarization.}
    \label{framework}
    \vspace{-8pt}
\end{figure*}

\section{Method}

\subsection{Overview}

\cref{framework} illustrates the pipeline of our Fully Aligned Network (FAN). Given an image and a descriptive language expression, a vision encoder and a language encoder extract visual and linguistic features. The image is encoded into hierarchical features $f_v$, and the text into fine-grained word embeddings $f_w$ and coarse-grained sentence embeddings $f_s$. A multi-scale activation module fuses these features to highlight the referent region and reduce background noise. 
Subsequently, the model embeds these features into a joint space, updating both of them with attention mechanisms in vision-to-language and language-to-vision decoders. Finally, the target region is isolated from the background using matrix multiplication.

\subsection{Image and Language Encoding}
For the input image $I \in \mathbb{R}^{H\times W\times 3}$, a pyramidal vision encoder extracts hierarchical features $f_v^i \in \mathbb{R}^{\frac{H}{2^{i}}\times \frac{W}{2^{i}}\times C_v^i}$, $i \in$ [2,3,4,5]. Here, $H$ and $W$ denote the height and width of the image, and $C$ denotes the channel dimension.

For the input text $L \in \mathbb{R}^{l}$, a transformer-based text encoder~\cite{CLIP, bert} encodes it into a word embedding $f_w \in \mathbb{R}^{l\times C_t}$ and a sentence embedding $f_s \in \mathbb{R}^{1\times C_t}$, where $l$ is the length of the text. The sentence embedding $f_s$ represents the overall characteristics of the target object, while the word embedding $f_w$ provides detailed information for precise segmentation.

\subsection{Activation Module}
We use a multi-scale activation module to preliminarily activate visual features with word embeddings $f_w$, highlighting the referred region. This reduces the background pixel influence on later alignment, aiding in the updating of linguistic and visual features. Our exploration showed that a multi-head cross-attention layer suffices for this activation.

The module takes word feature $f_w$ and hierarchical vision feature $f_v^i$, $i \in$ [2,3,4,5] as input. For the i-th scale, the visual feature $f_v^i$ serves as the query, and the word vector $f_w$ as the key and value. The process involves projecting input features to the corresponding space, applying multi-head attention to these projections, and then generating the activated cross-modal features $f_c^i$.

\subsection{Vision-to-Language Decoder}
We use the Vision-to-Language Decoder and Language-to-Vision Decoder to align visual and linguistic embeddings in a shared space.
The Vision-to-Language Decoder (V2L) takes an FPN-like architecture with a cross-modal alignment module. The Feature Pyramid Network (FPN)~\cite{fpn}, often used in object detection and segmentation, fuses multi-scale information and upsamples output features. We input multi-scale activated vision features $f_c$ with strides from 4$\times$ to 32$\times$. It outputs decoded 4$\times$ features. Fusion is performed from $f^5$ to $f^2$, and $f^2$ is 4$\times$ downsampled.

\begin{figure}[t]
    \centering
    \footnotesize
    \includegraphics[width=0.6\linewidth]{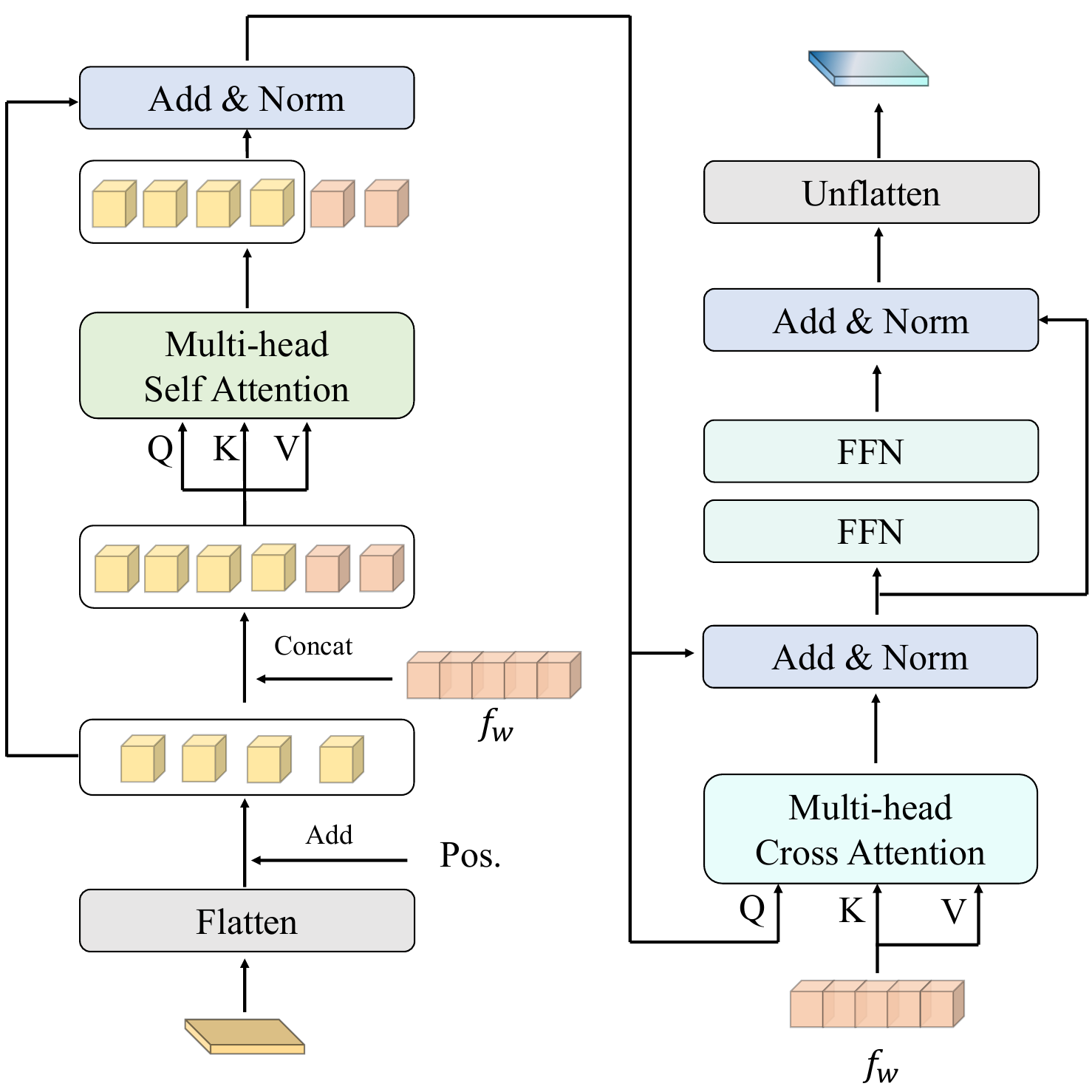}
    \caption{The structure of the Vision Projection Module (VPM).}
    \label{cross_modal-fpn}
    \vspace{-12pt}
\end{figure}

Unlike vanilla FPN, our V2L decoder integrates linguistic guidance into visual features using a Vision Projection Module (VPM) before multi-scale fusion, aiding in transferring visual features into a multi-modal space. The VPM structure (see \cref{cross_modal-fpn}) includes multi-modal self-attention and cross-attention layers. For the i-th level feature, we flatten it along the spatial dimension, add fixed positional embeddings~\cite{detr}, and concatenate the flattened tokens with word features $f_w$ to form multi-modal tokens. A multi-head self attention layer is applied to these tokens to extract relevant information and only vision tokens are selected for later cross-attention alignment.

This process allows the model to integrate information from both modalities while modeling the shared space. Fused vision tokens then serve as the query, and word embeddings $f_w$ as key and value for multi-head cross attention, aiding in locating the target object. Finally, the i-th level aligned vision features are output after residual connection and FFN~\cite{transformer} layers.

\subsection{Language-to-Vision Decoder}

For referring image segmentation, a common method involves fusing language embeddings with visual features and using the activated features for segmentation. However, this method does not fully utilize the representational ability of linguistic features. Unrestricted language expression can be ambiguous, especially in challenging scenes where language alone cannot clearly express the target object. For instance, the term ``pink" is vague until combined with an image context, such as a picture of two people, one wearing a pink dress, making ``pink dress" more informative.
Even if the description is detailed, it is given by humans based on their prior knowledge.
Due to differences in knowledge domain, models may not understand given descriptions well. 
This is somewhat similar to the recent research of prompt mechanism, which finds that learnable prompt embeddings work better than prompt defined by humans based on their own knowledge frameworks. 
Inspired by CLIP, which jointly learns visual and textual spaces, we use a Language-to-Vision Decoder (L2V) to align linguistic features to a multi-modal common space. By aligning linguistic features with the visual space, the output textual embedding becomes more perceptive to image content, providing a more informative description that better identifies the target object and distinguishes it from others in the image.

\subsection{Discussion of Framework and Principles}

% Our FAN adheres to the proposed cross-modal alignment principles.
% Here is a brief explanation.
% The activation module corresponds to the encoding interaction principle and highlights the referring region.
% % The preliminary interaction is somewhat like coding interaction.
% % LAVT~\cite{lavt} and EFN~\cite{efn}
% Different from LAVT~\cite{lavt} and EFN~\cite{efn} that perform interaction inside the visual backbone, we perform encoding interaction on the output feature maps of the backbone.
% The advantage of this approach is that it does not break the relationship between the pre-trained weights of the backbone, which allows us to take full advantage of the excellent pre-trained models such as CLIP~\cite{CLIP}.
% Besides, both the activation module and vision projection module take the hierarchical visual features as input, which corresponds to the multi-scale interaction principle.
% Under the guidance of the bidirectional interaction principle, we jointly update the visual and textual embeddings in the vision-to-language decoder and language-to-vision decoder to build the multi-modal common space.
% As for the coarse and fine-grained interaction principle, we utilize both fine-grained word embeddings $f_w$ and coarse-grained sentence embeddings $f_s$ for the V2L decoder and L2V decoder, respectively.
% This approach facilitates the use of detailed and holistic linguistic information to perceive the target object.
% Experiment results in \cref{tab:principle} also demonstrate the rationality and effectiveness of the above principles.

Our FAN adheres to the proposed cross-modal alignment principles. The activation module corresponds to the \textit{encoding interaction principle}, highlighting the referring region. Unlike LAVT~\cite{lavt} and EFN~\cite{efn}, which perform interaction within the visual backbone, we perform encoding interaction on the output feature maps. This preserves the pre-trained weights of the backbone, leveraging models like CLIP~\cite{CLIP}.Besides, 
both the activation module and vision projection module use hierarchical visual features, adhering to the \textit{multi-scale interaction principle}. Guided by the \textit{bidirectional interaction principle}, we update visual and textual embeddings in the vision-to-language and language-to-vision decoders to create a multi-modal common space. For the \textit{coarse and fine-grained interaction principle}, we use fine-grained word embeddings $f_w$ and coarse-grained sentence embeddings $f_s$ in the V2L and L2V decoders, respectively. This enables the use of detailed and holistic linguistic information to identify the target object.
Experiment results in \cref{tab:principle} demonstrate the validity and effectiveness of these principles.

\section{Experiment}

\input{Tables/main_results.tex}

\subsection{Datasets and Metrics.}

We used the following datasets: RefCOCO~\cite{refcoco}, derived from MSCOCO~\cite{mscoco}, is a key dataset for image segmentation and visual grounding, divided into training, validation, and test sets. RefCOCO\textbf{+}~\cite{refcoco} excludes certain location words and follows a similar split. G-Ref~\cite{grefcoco} features longer expressions with more location and appearance words, collected from Amazon Mechanical Turk.

For metrics, we use IoU and Precision$@$X~\cite{cris,vlt,rrn}, where IoU measures segmentation accuracy and Precision$@$X evaluates the location ability at various IoU thresholds.

\subsection{Implementation Details}
The model is implemented in Pytorch~\cite{pytorch}. Following \cite{cris}, we initialize the vision and language encoders with CLIP-ResNet50~\cite{CLIP} by default. We also experiment with other vision encoders like DeepLabV3~\cite{deeplabv3} pretrained ResNet101 and ImageNet~\cite{imagenet} pretrained Swin-B~\cite{swin} for fair comparison, with results shown in \cref{result}.
The Language-to-Vision decoder includes 6 transformer decoder layers, each with 8 heads and a feed-forward hidden dimension of 2048. The model is optimized using cross-entropy and dice loss. Considering extra [SOS] and [EOS] tokens, the maximum sentence length is 17 for RefCOCO~\cite{refcoco} and RefCOCO+~\cite{refcoco}, and 22 for G-Ref~\cite{grefcoco}. Input images are resized to 416 × 416.
We train the model with the Adam~\cite{adam} optimizer for 50 epochs on 8 Tesla V100 GPUs with a batch size of 64, taking about 7 hours. The initial learning rate is 0.0001, reduced by a factor of 0.1 at epoch 35. A smaller learning rate (scaling factor of 0.1) is set for the backbone.

For inference, the output mask is upsampled to the input image size by bilinear interpolation. Following~\cite{cris}, we binarize the prediction masks with a 0.35 threshold and do not use other post-processing operations.

\subsection{Comparison with State-of-the-arts}

In \cref{result}, we compare our FAN with previous state-of-the-art methods on the popular datasets RefCOCO, RefCOCO+, and G-Ref using the IoU metric. To enhance clarity, results using the same visual backbone are marked with the same color.
Our FAN achieves the best performance across all datasets. With the Swin-B backbone, FAN exceeds the previous SOTA method LAVT by 2\%. On the challenging G-Ref dataset, the margin extends to 4\% (\textbf{65.28} vs. \textbf{61.24}). Using the CLIP backbone, FAN surpasses previous methods significantly. Additionally, our model with ResNet-101 outperforms previous approaches using DarkNet and ViT. Notably, FAN with the CLIP-ResNet50 backbone even surpasses LAVT using Swin-B on some datasets, such as \textbf{62.83} vs. \textbf{62.14} on the RefCOCO+ val set.
These results demonstrate that our FAN, through effective alignment principles and simple attention operations, establishes a well-aligned vision-and-language common space, enhancing language-guided segmentation performance and simplifying the overall pipeline.

\input{Tables/alignment_principles.tex}

\subsection{Ablation Study}

\paragraph{Interaction Principles.}
\cref{tab:principle} demonstrates the importance of various types of interaction. Bidirectional Interaction enhances linguistic embeddings by integrating high-level visual information (row 1 \textit{vs} row 2). Multi-scale Interaction, which fuses linguistic and visual features at various scales, ensures segmentation accuracy and superior multi-modal understanding, with performance decreasing when fusion is limited to the highest level (row 3 \textit{vs} row 4). Encoding Interaction, involving preliminary activation of visual features, is crucial for coarse localization and minimizing background interference, with a 3\% performance drop observed without the Activation Module (row 4 \textit{vs} row 5). Lastly, Coarse and Fine-grained Interaction, utilizing both sentence-level and word-level features, provides better linguistic guidance than using sentence features alone (row 5 \textit{vs} row 6).

\input{Tables/some_results.tex}

\paragraph{Structure of Language-to-Vision Decoder.}

\cref{tab:some} shows that the number of transformer decoder layers has minimal impact on results, with one layer achieving 71.38 IoU, highlighting the lightweight nature of our FAN. Besides, using a transformer encoder is unnecessary since preliminary activation provides sufficient target objects. Our default setting uses no encoder layer and 6 decoder layers.

\paragraph{Structure of Vision Projection Module.}

The results of the ablation experiments summarized in \cref{tab:some} demonstrate that the Vision Projection Module's structure, which adopts a transformer decoder layer approach, is superior when integrating textual guidance into visual features through concatenation in the self-attention section followed by multi-modal information fusion via cross-attention, compared to using cross-attention alone.

\section{Conclusion}
In this paper, we address the referring image segmentation task by fully cross-modal alignment with eleborate attention mechanism.
We explicitly propose four interaction principles for aligning visual and textual information: encoding interaction, multi-scale interaction, coarse and fine-grained interaction, and bidirectional interaction.
Guided by the interaction principles, we propose a simple yet strong Fully Aligned Network (FAN), which achieves state-of-the-art performance on prevalent RIS benchmarks.

%%%%%%%%% REFERENCES
{\small
\bibliographystyle{IEEEtran.bst}
\bibliography{egbib}
}

\end{document}

%% file: Tables/main_results.tex
\begin{table*}[ht]
    \caption{Comparison with state-of-the-art methods in terms of the IoU metric on three popular benchmarks. We have experimented different visual backbone to perform fair comparison with other methods. To show the comparison more clearly, we mark the results of same level backbone with same color. Best viewed in color.}
    \centering
    \small
    % \scriptsize
    \setlength{\tabcolsep}{3.2mm}{\begin{tabular}{l|l|c|c|c|c|c|c|c|c}
      \toprule[1pt]
      \multirow{2}{*}{Method} &
      Vision &
      \multicolumn{3}{c|}{RefCOCO}  & \multicolumn{3}{c|}{RefCOCO+} & \multicolumn{2}{c}{G-Ref} \\
      \cline{3-10}
                                    & Backbone & val   & test A & test B & val & test A & test B & val & test   \\
      \hline
      % DMN~\cite{dmn} & ResNet101 & 49.78 & 54.83 & 45.13 & 38.88 & 44.22 & 32.29 & -     & -      \\
      % RRN~\cite{rrn}          & ResNet101 & 55.33 & 57.26 & 53.93 & 39.75 & 42.15 & 36.11 & -     & -      \\
      % MAttNet~\cite{mattnet}  & ResNet101 & 56.51 & 62.37 & 51.70 & 46.67 & 52.39 & 40.08 & 47.64 & 48.61      \\
      % CMSA~\cite{cmsa}       & ResNet101 & 58.32 & 60.61 & 55.09 & 43.76 & 47.60 & 37.89 & -     & - \\
      CAC~\cite{cac}        & ResNet101 & 58.90 & 61.77 & 53.81 & -     & -     & -       & 46.37 & 46.95  \\
      STEP~\cite{step}       & ResNet101 & 60.04 & 63.46 & 57.97 & 48.19 & 52.33 & 40.41 & -     & -      \\
      BRINet~\cite{brinet}    & ResNet101 & 60.98 & 62.99 & 59.21 & 48.17 & 52.32 & 42.11 & -     & -     \\
      CMPC~\cite{cmpc}     & ResNet101 & 61.36 & 64.53 & 59.64 & 49.56 & 53.44 & 43.23 & -     & -\\
      LSCM~\cite{lscm} & ResNet101 & 61.47 & 64.99 & 59.55 & 49.34 & 53.12 & 43.50 & -     & -\\
      CMPC+~\cite{cmpc+}      & ResNet101 & 62.47 & 65.08 & 60.82 & 50.25 & 54.04 & 43.47 & -     & -   \\
      MCN~\cite{mcn}       & DarkNet53 & 62.44 & 64.20 & 59.71 & 50.62 & 54.99 & 44.69 & 49.22 & 49.40 \\
      EFN~\cite{efn}                & ResNet101 & 62.76 & 65.69 & 59.67 & 51.50 & 55.24 & 43.01 & - & - \\
      BUSNet~\cite{busnet}          & ResNet101 & 63.27 & 66.41 & 61.39 & 51.76 & 56.87 & 44.13 & - & - \\
      CGAN~\cite{cgan}    & DarkNet53 & 64.86 & 68.04 & 62.07 & 51.03 & 55.51 & 44.06 & 51.01 & 51.69 \\
      LTS~\cite{lts}   & DarkNet53 & 65.43 & 67.76 & 63.08 & 54.21 & 58.32 & 48.02 & 54.40 & 54.25\\
      VLT~\cite{vlt}      & DarkNet56 & 65.65 & 68.29 & 62.73 & 55.50 & 59.20 & 49.36 & 52.99 & 56.65 \\
      ResTR~\cite{restr} & ViT-B & 67.22 & 69.30 & 64.45 & 55.78 & 60.44 & 48.27 & 54.48 & - \\
      CRIS~\cite{cris} & CLIP-ResNet50 & \textcolor{blue}{69.52} & \textcolor{blue}{72.72} & \textcolor{blue}{64.70} & \textcolor{blue}{61.39} & \textcolor{blue}{67.10} & \textcolor{blue}{52.48} & \textcolor{blue}{59.35} & \textcolor{blue}{59.39} \\
      LAVT~\cite{lavt} & Swin-B & \textcolor{red}{72.73} & \textcolor{red}{75.82} & \textcolor{red}{68.79} & \textcolor{red}{62.14} & \textcolor{red}{68.38} & \textcolor{red}{55.10} & \textcolor{red}{61.24} & \textcolor{red}{62.09} \\
      \hline
      FAN (Ours) & ResNet101 & \textbf{69.42} & \textbf{71.25} & \textbf{64.82} & \textbf{58.84} & \textbf{62.46} & \textbf{51.55} & \textbf{58.75} & \textbf{58.93}\\
      FAN (Ours) & CLIP-ResNet50 & \textcolor{blue}{\textbf{71.67}} & \textcolor{blue}{\textbf{74.58}} & \textcolor{blue}{\textbf{66.55}} & \textcolor{blue}{\textbf{62.83}} & \textcolor{blue}{\textbf{68.95}} & \textcolor{blue}{\textbf{53.15}} & \textcolor{blue}{\textbf{60.49}} & \textcolor{blue}{\textbf{61.32}}\\
      FAN (Ours) & Swin-B & \textcolor{red}{\textbf{74.06}} & \textcolor{red}{\textbf{75.97}} & \textcolor{red}{\textbf{70.84}} & \textcolor{red}{\textbf{64.14}} & \textcolor{red}{\textbf{69.08}} & \textcolor{red}{\textbf{58.53}} & \textcolor{red}{\textbf{65.28}} & \textcolor{red}{\textbf{65.51}}\\
      \bottomrule[1pt]
   \end{tabular}}
   
   \label{result}
\end{table*}

%% file: Tables/alignment_principles.tex
% \begin{table}[t]
%    \centering
%    \renewcommand\arraystretch{1.2}
%    \resizebox{\columnwidth}{!}{\begin{tabular}{l|c|c|c|c}
%       \toprule[1pt]
%       & IoU & P@0.5 & P@0.7 & P@0.9\\
%       \bottomrule[1pt]
%       \multicolumn{5}{l}{ \textit{(a) Encoding Alignment}} \\
%       \toprule[1pt]
%       W/ Preliminary Fusion Module & 71.67 & 82.80 & 71.13 & 21.91  \\
%       w/o Preliminary Fusion Module & - & - & - & -\\
%       \bottomrule[1pt]
%       \multicolumn{5}{l}{ \textit{(b) Multi-scale Alignment}} \\
%       \toprule[1pt]
%       Multi-scale Fusion  & 71.67 & 82.80 & 71.13 & 21.91  \\
%       Single-scale Fusion  & - & - & - & - \\
%       \bottomrule[1pt]
%       \multicolumn{5}{l}{ \textit{(c) Coarse and Fine-grained Alignment}} \\
%       \toprule[1pt]
%       W/ Word Embedding  & 71.67 & 82.80 & 71.13 & 21.91 \\
%       W/o Word Embedding & - & - & - & -  \\
%       \bottomrule[1pt]
%       \multicolumn{5}{l}{ \textit{(d) Bidirectional Alignment}} \\
%       \toprule[1pt]
%       W/ Language Decoder & 71.67 & 82.80 & 71.13 & 21.91\\
%       W/o Language Decoder & 71.18 & 82.26 & 70.86 & 21.10  \\
%       \bottomrule[1pt]
%    \end{tabular}}
%    \caption{Ablation studies about the proposed interaction principles on the RefCOCO validation set. }
%    \label{tab:principle}
% \end{table}

\begin{table}[t]
    \caption{Ablation studies about the proposed interaction principles on the RefCOCO validation set.}
    \centering
    % \small
    \small
    \renewcommand\arraystretch{1.1}
    \setlength\tabcolsep{3pt}
    \begin{tabular}{l|c|c|c}
        \toprule[1pt]
        Model & IoU & P@0.5 & P@0.9\\
        \hline
        Simple Baseline  &59.30 &66.49 &10.85  \\
        + Language-to-Vision Decoder	&64.25 &72.88 &16.28 \\
        + Single-Scale Vision Projection Module	&67.97 &77.94 &18.56 \\
        + Multi-Scale Vision Projection Module	&68.72 &79.17 &19.65 \\
        + Activation Module	&\textbf{71.67} &\textbf{82.80} &\textbf{21.91} \\
        Only utilize sentence embedding	&69.88 &81.12 &19.84 \\
        \bottomrule[1pt]
    \end{tabular}
    \vspace{-10pt}
    \label{tab:principle}
    \end{table}

%% file: Tables/some_results.tex
\begin{table}[t]
    \caption{Experiments about structure of Language-to-Vision Decoder. The vision encoder used is CLIP-ResNet50~\cite{CLIP}.}
   \centering
   \small
   % \scriptsize
  % \renewcommand\arraystretch{1.1}
  \setlength{\tabcolsep}{4pt}
   \begin{tabular}{l|c|c|c}
      \toprule[1pt]
      & IoU & P@0.5  & P@0.9\\
      \hline
      % \multicolumn{4}{l}{ \textit{(a) Leanrnable Query}} \\
      % \hline
      % Sentence Embedding & 71.67 & 82.80 & 21.91  \\
      % + Learnable Query & 70.47 & 81.17 & 20.80 \\
      % \hline
      \rowcolor{gray!18}\multicolumn{4}{l}{ \textit{(a) Structure of Language-to-Vision Decoder}} \\
      \hline
      1 Decoder Layer  & 71.38 & 82.36 & 21.40 \\
      3 Decoder Layers  & 71.45 & 82.43 & 21.33   \\
      6 Decoder Layers  & 71.67 & 82.80 & 21.91   \\
      + Encoder Layers  & 71.67 & 82.92 & 21.93   \\
      \hline
      \rowcolor{gray!18}\multicolumn{4}{l}{ \textit{(b) Structure of Vision Projection Module}} \\
      \hline
      % Only Self-Attention Fusion & 66.97 & 76.73 & 18.58  \\
      % Only  Cross-Attention Fusion & 67.00 & 76.68 & 18.10   \\
      % Both Self and Cross-Attention Fusion & 68.13 & 78.37 & 18.8   \\
      %%%%%%%%%%%%%%%%%%%%%%%%%%%%%%%%%% above is no encoding fusion, below is all
      Only  Cross-Attention Fusion & 68.55 & 78.64 & 19.38   \\
      Both Self and Cross-Attention Fusion & 71.67 & 82.92 & 21.93   \\
      % \hline
      % \multicolumn{4}{l}{ \textit{(d) Disadvantage of Fusion Inside Backbone}} \\
      % \hline
      % LAVT~\cite{lavt} with Swin-B & 72.73 & 84.46 & 34.3  \\
      % LAVT~\cite{lavt} with CLIP & - & - & -   \\
      % \hline
      % Ours with Swin-B & 74.06 & 85.82 & 24.77  \\
      % Ours with CLIP & 71.67 & 82.80 & 21.91  \\
      \bottomrule[1pt]
   \end{tabular}
   
   \label{tab:some}
   \vspace{-15pt}
\end{table}